\documentclass[conference,compsoc]{IEEEtran}
\ifCLASSOPTIONcompsoc
  \usepackage[nocompress]{cite}
\else
  \usepackage{cite}
\fi
\ifCLASSINFOpdf
   \usepackage[pdftex]{graphicx}
   \graphicspath{{./}}
   \DeclareGraphicsExtensions{.pdf,.jpg,.jpeg,.png}
\else
\fi
\usepackage{hyperref}

\usepackage{multirow}

\begin{document}
%
\title{Accelerating Translational Image Registration for HDR Images on GPU}

\author{\IEEEauthorblockN{Kadir Cenk Alpay *}
\IEEEauthorblockA{Department of Computer Engineering\\
Middle East Technical University\\
kadircenk@ceng.metu.edu.tr}
\and
\IEEEauthorblockN{Kadir Berkay Aydemir *}
\IEEEauthorblockA{Graduate School of Informatics\\
Middle East Technical University\\
berkay.aydemir@metu.edu.tr}
\and
\IEEEauthorblockN{Alptekin Temizel}
\IEEEauthorblockA{Graduate School of Informatics\\
Middle East Technical University\\
atemizel@metu.edu.tr}}


%

\IEEEspecialpapernotice{(Submitted for Consideration for Publication in High Performance Computing Conference 2020)}

\maketitle

\begin{abstract}
High Dynamic Range (HDR) images are generated using multiple exposures of a scene. When a hand-held camera is used to capture a static scene, these images need to be aligned by globally shifting each image in both dimensions. For a fast and robust alignment, the shift amount is commonly calculated using Median Threshold Bitmaps (MTB) and creating an image pyramid. In this study, we optimize these computations using a parallel processing approach utilizing GPU. Experimental evaluation shows that the proposed implementation achieves a speed-up of up to 6.24 times over the baseline multi-threaded CPU implementation on the alignment of one image pair. The source code is available at \textit{\url{https://github.com/kadircenk/WardMTBCuda}}\\

\end{abstract}

\noindent $*$ Kadir Cenk Alpay and Kadir Berkay Aydemir contributed equally to this work.


%
\IEEEpeerreviewmaketitle

\section{Introduction}\label{introduction}
\begin{figure}[!t] \par
\centering
\frame{\includegraphics[width=0.478\textwidth]{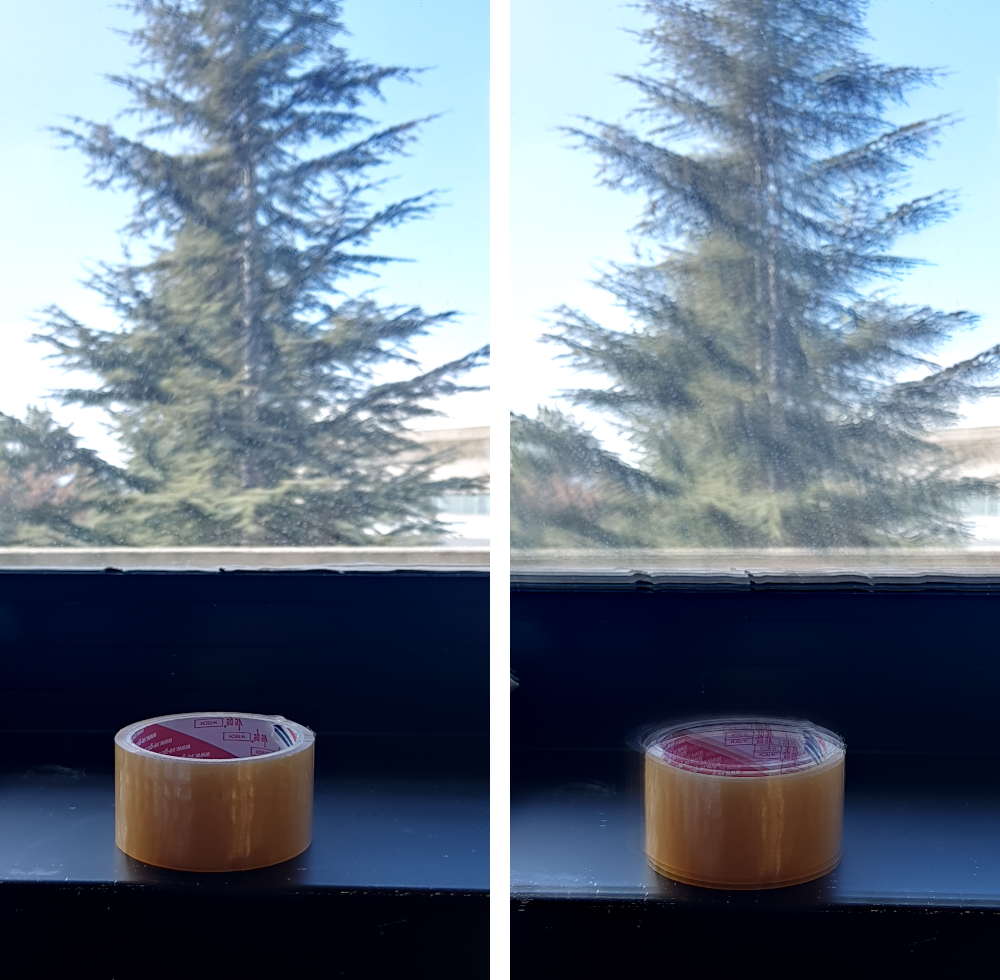}}
\caption{Blurry artifacts occur when non-aligned images are used to create an HDR image (right). Artifacts are eliminated when images are aligned before merging (left).}
\label{fig:blurryartifact}
\end{figure}

In several applications, a High Dynamic Range (HDR) image is obtained by fusing multiple images of the same scene taken with different exposures by a standard camera \cite{debevec} \cite{mertens}. By doing so, the resulting HDR image includes scene detail from well-exposed regions of each image. Using a handheld camera to capture a static scene with multiple exposure levels results in misaligned images due to shaking of the camera. Using a tripod to fix the position of the camera reduces the amount of misalignment; however, this brings a significant restriction and may not be feasible at all shooting conditions. In addition, small shifts may still occur due handling of the camera to adjust its parameters or to shoot the photo. This results in a blurry HDR image when multiple misaligned images are merged (Figure \ref{fig:blurryartifact}).\par
\par
The main steps of an HDR  pipeline are: (\textit{i}) acquisition of multiple exposure images of a scene, (\textit{ii}) aligning them to each other, (\textit{iii}) deriving the camera response function, and (\textit{iv}) computing the final HDR image \cite{reinhard}. The alignment phase includes computationally expensive 2D image manipulation operations. This study focuses on this alignment phase to decrease the time cost of the overall HDR pipeline.\par
Traditional image alignment algorithms are not suitable for the alignment of images of the same static scene with varying exposures. Hence, specialized algorithms were proposed for HDR imaging pipeline. A popular algorithm amongst these is the translational image registration method that uses Median Threshold Bitmaps (MTB) \cite{ward}. Utilization of MTBs alleviates the effect of varying exposure times and brings the images to a common data format. This way, MTBs of the images can be shifted and error-tested to automatically calculate the globally optimal shift amount.\par
In this paper, we focus on speeding-up an MTB based alignment algorithm \cite{ward}, and image manipulation routines by leveraging the highly parallel nature of GPUs. For this purpose, we optimize the following operations on the GPU: converting multiple exposures into gray-scale, downsampling, shifting, histogram calculations, creation of bitmaps, pixel-wise AND and XOR operations and erroneous pixel counting. \par
The remainder of this paper is organized as follows: In Section \ref{relatedwork}, we provide a summary of related work and their limitations. In Section \ref{proposedmethod}, we describe the reference method and proposed optimizations on the GPU. In Section \ref{experimentalevaluation}, we provide the experimental evaluation and discussion of the results. In Section \ref{conclusionandfuturework}, we provide the conclusion and future work ideas.

\section{Related Work}\label{relatedwork}

The method introduced in \cite{ward} for registration of handheld exposures of a static scene for HDR imaging has become a landmark work due to its robustness to high exposure differences. The fundamental idea in this work is making use of MTBs to eliminate the brightness difference caused by different exposure times. This idea has been found to be highly effective and many follow-up image registration techniques have been introduced to enable registration of images captured with different exposure times.\par
Ward \cite{ward} claimed that only 10\% of the test scenes requires an additional rotational alignment along with the translational alignment. While this work did not implement rotational alignment; it suggests splitting the image into quadrants and then applying the algorithm on each quadrant separately to detect if rotational alignment is needed. This study was further extended by implementation of an additional rotational alignment step, along with the handling of small movements of in-scene objects \cite{grosch}. Moreover, graphical hardware was used to calculate XOR result of two MTBs by first converting them into textures, and then running a fragment code on them. Another method, extending for rotational alignment, integrates rotation into error tests. By rotating the image with steps of 0.5 degree, pivoting at the center \cite{jacobs}, this approach finds the optimal rotation angle together with the translation offset.\par
Akyüz \cite{akyuz} uses correlation maps to find a suitable translational alignment offset, as correlation maps are not dependent on exposure time. Khan et al. \cite{khan} iteratively weight the pixels to detect their contribution to the final HDR image to remove blurry artifacts. Sen et al. \cite{sen} apply a joint optimization equation which solves the alignment and the HDR reconstruction problems simultaneously. Zimmer et al. \cite{zimmer} use an energy‐based optic flow approach which can deal with different exposure timings to perform the alignment with in-scene object movement.\par

\begin{figure}[!t]
\centering
\frame{\includegraphics[width=0.478\textwidth]{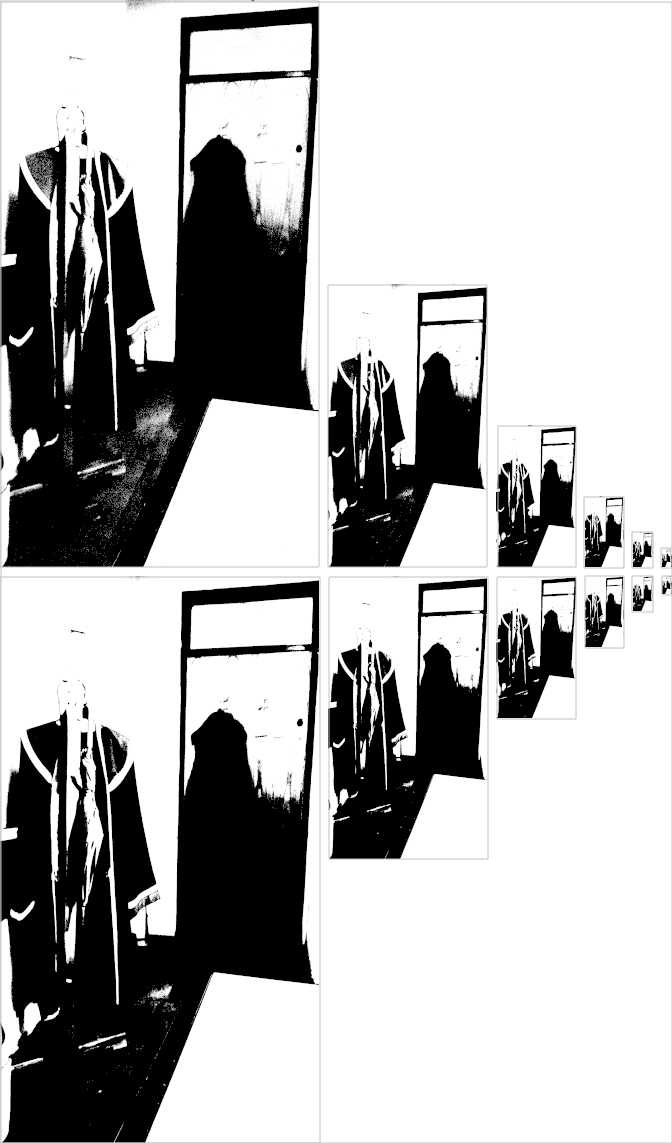}}
\caption{A 6-level image pyramid of MTBs of two exposures in comparison. Sizes are shown in relative scale, halved in both dimensions at each level.}
\label{fig:imagepyramid}
\end{figure}

Guthier et al. \cite{guthier} propose parallelization of the histogram-based exposure registration using CUDA where parts of the MTB algorithm are run on the GPU. They register exposure pairs based on \cite{ward}, but different to that, they extract row and column histograms from each MTB of the image pair and apply Normalized Cross Correlation on the corresponding histograms to calculate the shift offset in each dimension, without utilizing an image pyramid. Different from \cite{guthier}, the implementation proposed in this paper adheres to the original algorithm \cite{ward}, and also run all image-related calculations on the GPU, eliminating costly GPU-CPU memory copy operations. In addition, as the code is run iteratively for each image pair, the copy operations bring higher overhead as a some images need to be copied twice (once to compare with the previous image and once to compare with the next image)  when input image count is higher than two. The implementation proposed in this paper is designed to copy all the input images to the GPU memory once, avoiding multiple copies.\par
As both \cite{grosch} and \cite{jacobs} are based on MTB, the proposed implementation can also be used to speed-up their translational alignment. In the reference implementation, CPU processes the 2D input image stack in a sequential manner or in the multi-threaded case, it works by assigning parts of the image to different CPU threads. On the other hand, the GPU implementation can assign each pixel to a different GPU thread to massively parallelize the algorithm. A further higher level parallelism can also be introduced by assigning each image to a different CUDA stream and run them in parallel.

\section{Reference Method and Parallelization}\label{proposedmethod}

In this section, we first describe the MTB based reference algorithm and then describe the proposed optimization of this method.
\subsection{MTB Based Image Registration Algorithm}
MTB based algorithm \cite{ward} first converts the exposures into gray-scale and calculates the median pixel value of each exposure over its computed histogram. Then, a median thresholding is applied to binarize these images into MTBs. This step replaces the pixels having a value less than or equal to the median with a value of zero and the others with a value of one in MTBs. Shift tests are done by applying pixel-wise XOR operation on the MTBs of the image pair in comparison. Alignment error is then found by counting the non-zero values in the XOR result.\par
A brute-force approach to align two MTBs would go through each possible shift offset to find out the optimal one, which requires $n^2$ alignment tests for an $n$ pixel shift in both dimensions. A computationally more efficient alternative approach is using image pyramids. In this approach, a 6-level image pyramid is created by down-sampling the gray-scale images by halving at both dimensions and creating the MTBs at each level \cite{ward} (Figure \ref{fig:imagepyramid}). Then starting from the deepest level, only $\pm$ 1  pixel shift tests are applied both horizontally and vertically. Multiplying the total shift offset by two before starting with the next level until all levels are done, each bit of the global shift offset is found in only 54 shift tests.\par

\subsection{Implementation and Parallelization}

\begin{figure}[!t]
\centering
\frame{\includegraphics[width=0.478\textwidth]{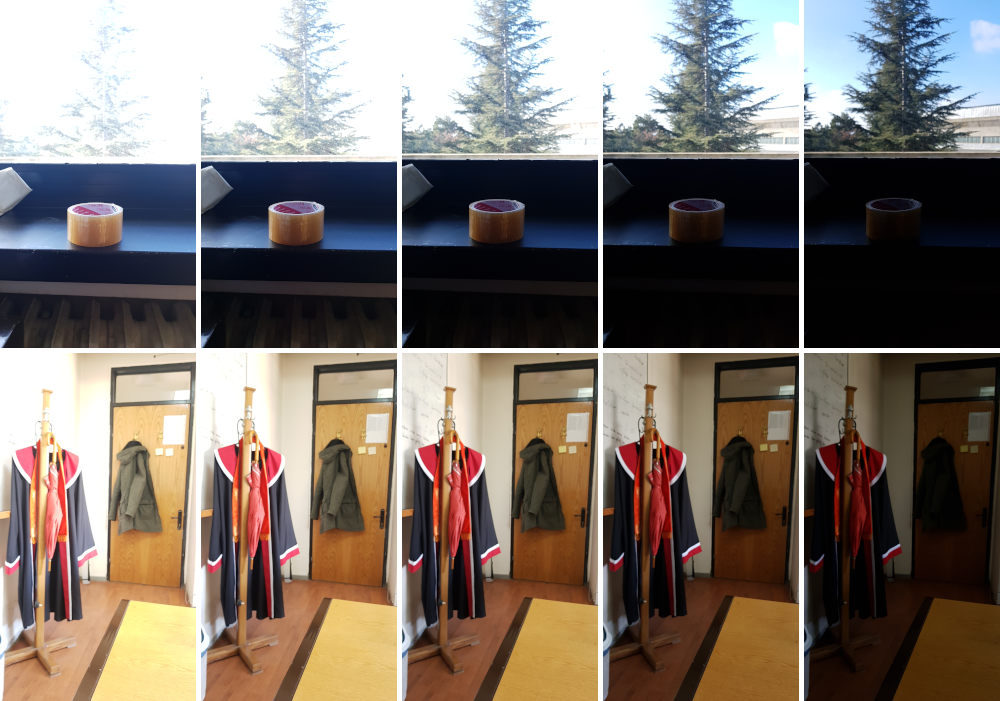}}
\caption{3.7 MP exposure sets. Each row shows a set of 5 decreasing exposure images of a different scene.}
\label{fig:dataset}
\end{figure}
\begin{figure}[!t]
\centering
\includegraphics[width=0.5\textwidth]{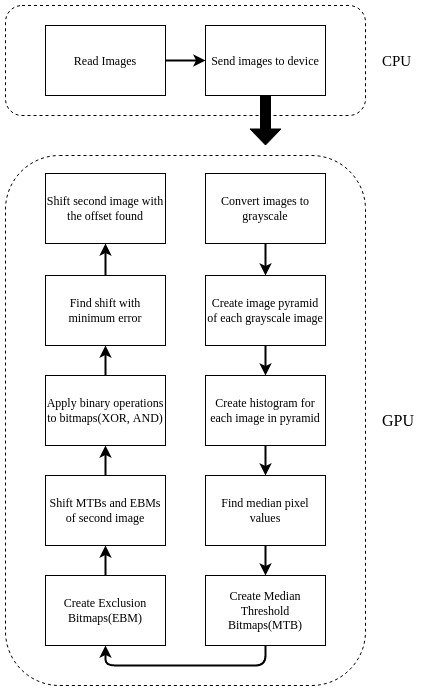}
\caption{Flow of the proposed parallel implementation of the MTB based method \cite{ward}. }
\label{fig:flow}
\end{figure}

In our approach (Figure \ref{fig:flow}), after the input images are read into the CPU memory, they are sent to the GPU memory and separate CUDA kernels are launched using one CUDA stream for each image to calculate the gray-scale images in parallel  \cite{cudastream}. Then, these gray-scale images are bound as CUDA textures and downsampled by launching linear interpolation downsampling kernels in all streams. This is done iteratively to create 5 levels of MTB pyramid. Using CUDA texture memory grants a speed-up by decreasing the neighbouring pixel access time on the GPU memory \cite{cudatexmem}.\par

\begin{table*}[!t]
\caption{Timing Results in ms.}
\label{tab:my-table}
\centering
\resizebox{\textwidth}{!}{%
\begin{tabular}{|c|c|r|r|r|r|r|r|}
\hline
\multirow{2}{*}{\textbf{Implementation}} & \multirow{2}{*}{\textbf{Configuration}} & \multicolumn{6}{c|}{\textbf{Input Image Count}}                                                                                                                                                                                  \\
                                         &                                         & \multicolumn{1}{c}{\textbf{2}}      & \multicolumn{1}{c}{\textbf{3}}      & \multicolumn{1}{c}{\textbf{4}}       & \multicolumn{1}{c}{\textbf{5}}       & \multicolumn{1}{c}{\textbf{6}}       & \multicolumn{1}{c|}{\textbf{7}} \\ \hline
\multirow{3}{*}{OpenCV}                  & CPU 1                                      & 68.05          & 135.62         & 200.73          & 273.82          & 332.15          & 385.56                          \\  \cline{2-8}
                                         & CPU 2                                      & 41.02          & 79.53          & 118.37          & 156.08          & 186.09          & 219.88                          \\  \cline{2-8}
                                         & CPU 3                                      & 34.14              & 64.48              & 95.13               & 127.28               & 161.53      & 190.45                \\ \hline \hline
\multirow{3}{*}{Ours}                    & GPU 1                                      & \textbf{45.44} & \textbf{85.43} & \textbf{124.15} & \textbf{165.70} & \textbf{208.55} & \textbf{252.76}                 \\  \cline{2-8}
                                         & GPU 2                                      & \textbf{13.38} & \textbf{24.98} & \textbf{36.08}  & \textbf{47.38}  & \textbf{58.96}  & \textbf{70.51}                  \\  \cline{2-8}
                                         & GPU 3                                      & \textbf{10.90} & \textbf{19.99} & \textbf{28.60}  & \textbf{38.09}  & \textbf{47.91}  & \textbf{57.21}                  \\ \hline
\end{tabular}%
}
\end{table*}

In the next step, pixel histogram calculation kernels are launched to calculate 256-bin histograms on all gray-scale images in the pyramid. The kernels use parallel reduction technique to speed-up the calculation \cite{cudaparallelreduction}. Then, another kernel is launched to calculate the median value of each histogram in the pyramid. Finally, MTBs of all images are created using the median value by another kernel.\par
The MTBs of two images to be compared are XORed to get the error bitmap of the current translational shift offset candidate. However, using the XOR result directly is prone to errors, as hard-cutting the pixel's MTB value right at the median is prone to create noise at the image area with pixels that are close to the median value. For this purpose, \textit{Exclusion Bitmaps (EB)} \cite{ward} are used. In EBs, each bit is set to one if that pixel's value is $\pm$ 4 of the median and to zero otherwise. Later, the EBs of both images in comparison are ANDed onto the XOR result to remove the noisy bits that would otherwise negatively affect the robustness of the algorithm.\par

Each white pixel in the  denoised XOR image accounts for an error and they need to be counted to find out the final error. This procedure is repeated by shifting the MTB of the second image to test the remaining shift offset candidates, at each level of the pyramid. By propagating the offset up to the top-most level of the pyramid, global offset is calculated. Lastly, for each image pair in the input stack, the second image is shifted using this offset to match the first image.\par

AND, XOR, error-counting and gray-scale image shifting functions have been implemented as separate CUDA kernels. These kernels are overlapped whenever possible by placing them into separate CUDA streams. Once the global offset is calculated, RGB image shifting kernel is used to apply this shift offset to each channel of the second image, aligning it with the reference.\par
The MTB algorithm is iterated to pairwise align the input images if there are more than two images. For instance, an image set with 5 images requires 4 iterations of the algorithm. All the images are sent to the GPU memory at the beginning of the execution and corresponding MTB pyramids are calculated once. So, no CPU-to-GPU memory copies are needed for the further iterations.\par

In the reference algorithm, one bit is used to hold each pixel’s MTB value and bit-wise operations are used to speed-up the CPU implementation \cite{ward}. The OpenCV implementation also uses this idea \cite{opencv}. 
However, converting each pixel of the bitmap image from one byte to one bit brings an overhead during the pre-processing phase of the algorithm. On the other hand, in GPU implementations, for such low arithmetic intensity operations, memory transfers are likely to be the bottleneck. So, in our approach, we used  uint8\_t (unsigned char) type, as GPU memory throughput is better when appropriate variable types are used in CUDA kernels. Due to this fact, we used byte-wise rather than bit-wise operations and eliminated the overhead that would be caused by byte-to-bit conversion. So, in our implementation, instead of bitmaps, we hold byte-maps on the GPU, where we set the pixel value to 255 instead of 1, and 0 otherwise. The source code of the implementation is available at \textit{\url{https://github.com/kadircenk/WardMTBCuda}}\par

\begin{table}
\centering
\caption{Speed-up of our implementation on one image pair.}
\label{tab:speeduptable}
\resizebox{0.48\textwidth}{!}{%
\begin{tabular}{c|r|r|r|}
\cline{2-4}
                                  & \multicolumn{1}{|c|}{\textbf{CPU 1}} & \multicolumn{1}{|c|}{\textbf{CPU 2}} & \multicolumn{1}{|c|}{\textbf{CPU 3}} \\ \hline
\multicolumn{1}{|c|}{\textbf{GPU 1}} & 1.50                            & 0.90                            & 0.75                             \\ \hline
\multicolumn{1}{|c|}{\textbf{GPU 2}} & 5.08                            & 3.06                            & 2.55                             \\ \hline
\multicolumn{1}{|c|}{\textbf{GPU 3}} & 6.24                            & 3.76                            & 3.13                             \\ \hline
\end{tabular}%
}
\end{table}

\section{Experimental Evaluation and Discussion}\label{experimentalevaluation}
We have evaluated the proposed implementation with different hardware configurations and two diverse image sets. OpenCV includes a multi-threaded and publicly available CPU implementation \cite{opencv} of the original MTB algorithm \cite{ward} and we use it as the baseline method to compare against.\par
All hardware configurations are running Ubuntu 18.04 LTS, and they are detailed as follows. CPU 1: 4-core 2.4GHz Intel® i7-4700HQ. CPU 2: 6-core 3.20GHz Intel® i7-8700. CPU 3: 8-core 3.6GHz Intel® Core™ i7-9700K. GPU 1: NVIDIA GeForce GTX 850M. GPU 2: NVIDIA GeForce GTX 1080Ti. GPU 3: NVIDIA GeForce RTX 2080Ti.\par
The test image set consist of 3.7 megapixel (MP) exposures (Figure \ref{fig:dataset}).
For all the results, we executed the codes 10 times and calculated the average run-time. Note that the run-time of the algorithm depends only on the image resolution and it is independent of the image content. While measuring the run-time, we excluded the reading/writing times of the images from/to the disk for both implementations. However, as our approach requires the images to be available in the GPU memory, CPU to GPU data transfer cost is included in the total run-time to compare on a fair-ground.\par
The algorithm needs at least a single pair of images and as more than 7 exposures in one input set is not common, we experiment with at most 7 images. We provide the run-times of both implementations by varying the number of input images from 2 to 7 in Table \ref{tab:my-table}. 
As both implementations iterate the MTB algorithm for each image pair, number of input images linearly increases the time cost.\par


Speed-up gains of our implementation over the CPU code on the aforementioned hardware are provided in Table \ref{tab:speeduptable}. The GPU implementation on a low-spec GPU board (GPU 1) cannot grant a speed-up over high-spec CPU models (CPU 2 and CPU 3). On the other hand when a high-spec GPU (GPU 3) is used it can provide a speed-up of 3.13 times against a high-spec CPU (CPU 3) and 3.76 and 6.24 times speed-up against CPU 2 and 1 respectively.\par
Grosch \cite{grosch} aligns five images each having 1024x768 resolution in 4 seconds; together with rotational alignment and object movement removal. Our study, on the other hand, focuses mainly on speeding-up the translational alignment phase. As our runtimes are at most in milliseconds for images with resolution 2560x1440, our work can also be integrated to speed-up this algorithm.\par

\section{Conclusions and Future Work}\label{conclusionandfuturework}
In this study, we have proposed a pipeline which makes use of GPU to translationally register the input stack of handheld exposures using the MTB alignment algorithm \cite{ward}. Experimental evaluations show that this multi-threaded and multi-stream implementation has a speed-up of up to 6.24 times compared to the baseline OpenCV multi-threaded MTB implementation on CPU. This speed-up facilitates integration of this algorithm into a real-time HDR video sequence pipeline on a lower cost GPU based system. 

The reference MTB alignment algorithm requires the scene to be static \cite{ward} and any in-scene object motion would have a negative effect on the performance by causing erroneous MTB values during alignment testing. As a future work, an object motion detection algorithm could be integrated into our work to enable the alignment of such scenes. In addition, the method could be extended with a basic rotational alignment step, which tests for the optimal rotation angle step by step together with the translation offset.\par

\vfill\break


\begin{thebibliography}{1}

\bibitem{debevec}
Debevec, P. E., \& Malik, J. (2008). Recovering high dynamic range radiance maps from photographs. In ACM SIGGRAPH 2008 classes (pp. 1-10).

\bibitem{mertens}
Mertens, T., Kautz, J., \& Van Reeth, F. (2009, March). Exposure fusion: A simple and practical alternative to high dynamic range photography. In Computer graphics forum (Vol. 28, No. 1, pp. 161-171). Oxford, UK: Blackwell Publishing Ltd.

\bibitem{reinhard}
Reinhard, E., Heidrich, W., Debevec, P., Pattanaik, S., Ward, G., \& Myszkowski, K. (2010). High dynamic range imaging: acquisition, display, and image-based lighting. Morgan Kaufmann.

\bibitem{ward}
Ward, G. (2003). Fast, robust image registration for compositing high dynamic range photographs from hand-held exposures. Journal of graphics tools, 8(2), 17-30.

\bibitem{grosch}
Grosch, T. (2006). Fast and robust high dynamic range image generation with camera and object movement. Vision, Modeling and Visualization, RWTH Aachen, 277284.

\bibitem{jacobs}
Jacobs, K., Loscos, C., \& Ward, G. (2008). Automatic high-dynamic range image generation for dynamic scenes. IEEE Computer Graphics and Applications, 28(2), 84-93.

\bibitem{akyuz}
Akyüz, A. O. (2011, April). Photographically Guided Alignment for HDR Images. In Eurographics (Areas Papers) (pp. 73-74).

\bibitem{khan}
Khan, E. A., Akyuz, A. O., \& Reinhard, E. (2006, October). Ghost removal in high dynamic range images. In 2006 International Conference on Image Processing (pp. 2005-2008). IEEE.

\bibitem{sen}
Sen, P., Kalantari, N. K., Yaesoubi, M., Darabi, S., Goldman, D. B., \& Shechtman, E. (2012). Robust patch-based hdr reconstruction of dynamic scenes. ACM Trans. Graph., 31(6), 203-1.

\bibitem{zimmer}
Zimmer, H., Bruhn, A., \& Weickert, J. (2011, April). Freehand HDR imaging of moving scenes with simultaneous resolution enhancement. In Computer Graphics Forum (Vol. 30, No. 2, pp. 405-414). Oxford, UK: Blackwell Publishing Ltd.

\bibitem{guthier}
Guthier, B., Kopf, S., Wichtlhuber, M., \& Effelsberg, W. (2012, April). Parallel algorithms for histogram-based image registration. In 2012 19th International Conference on Systems, Signals and Image Processing (IWSSIP) (pp. 172-175). IEEE.

\bibitem{opencv}
"cv::AlignMTB Class Reference", OpenCV. [Online]. Available: https://docs.opencv.org/3.4/d7/db6/classcv\_1\_1AlignMTB.html. [Accessed: 01-Jun-2020].

\bibitem{cudastream}
"Programming Guide :: CUDA Toolkit Documentation - Streams", NVIDIA. [Online]. Available: https://docs.nvidia.com/cuda/cuda-c-programming-guide/index.html\#streams. [Accessed: 02-Jun-2020].

\bibitem{cudatexmem}
"Programming Guide :: CUDA Toolkit Documentation - Texture and Surface Memory", NVIDIA. [Online]. Available: https://docs.nvidia.com/cuda/cuda-c-programming-guide/index.html\#texture-and-surface-memory. [Accessed: 02-Jun-2020].

\bibitem{cudaparallelreduction}
Harris, M., "Optimizing Parallel Reduction in CUDA", NVIDIA Developer Technology. [Online]. Available: https://developer.download.nvidia.com/assets/cuda/files/reduction.pdf. [Accessed: 02-Jun-2020].

\end{thebibliography}
\end{document}